\documentclass{article}
\usepackage{ijcai18}

\usepackage{times}
\usepackage{xcolor}
\usepackage{soul}
\usepackage[utf8]{inputenc}
\usepackage[small]{caption}
\usepackage{graphicx}
\usepackage{hyperref}
\usepackage{amsmath}
\usepackage{amssymb}
\usepackage{bm}
\usepackage{booktabs}

\title{Better and Faster: Knowledge Transfer from Multiple Self-supervised Learning Tasks via Graph Distillation for Video Classification}

\author{{Chenrui Zhang} and {Yuxin Peng}\thanks{Corresponding author.} \\ Institute of Computer Science and Technology, Peking University \\ Beijing 100871, China \\
pengyuxin@pku.edu.cn}

\begin{document}

\maketitle

\begin{abstract}
  Video representation learning is a vital problem for classification task. Recently, a promising unsupervised paradigm termed \textit{self-supervised learning} has emerged, which explores inherent supervisory signals implied in massive data for feature learning via solving auxiliary tasks. However, existing methods in this regard suffer from two limitations when extended to video classification. First, they focus only on a single task, whereas ignoring complementarity among different task-specific features and thus resulting in suboptimal video representation. Second, high computational and memory cost hinders their application in real-world scenarios. In this paper, we propose a graph-based distillation framework to address these problems: (1) We propose logits graph and representation graph to transfer knowledge from multiple self-supervised tasks, where the former distills classifier-level knowledge by solving a multi-distribution joint matching problem, and the latter distills internal feature knowledge from pairwise ensembled representations with tackling the challenge of heterogeneity among different features; (2) The proposal that adopts a teacher-student framework can reduce the redundancy of knowledge learnt from teachers dramatically, leading to a lighter student model that solves classification task more efficiently. Experimental results on 3 video datasets validate that our proposal not only helps learn better video representation but also compress model for faster inference.
\end{abstract}

\section{Introduction}

Video representation learning aims to capture discriminative features from video data and is a critical premise for classification problem. In the last decade, supervised methods have achieved remarkable success in a variety of areas, showing extraordinary performance on representation learning. However, heavy reliance on well-labeled data limits the scalability of these methods as building large-scale labeled datasets is time-consuming and costly. Furthermore, learning from manual annotations is inconsistent with biology, as living organisms develop their visual systems without the requirement for millions of semantic labels. Hence, there is a growing interest in unsupervised learning, and \textit{self-supervised learning}, one of the most promising unsupervised representation learning paradigms, is gaining momentum.

In contrast to supervised methods, self-supervised learning exploits structural information of the raw visual data as supervisory signals to yield transferable representation without manual annotations. Concretely, machine is asked to solve an auxiliary task by leveraging self-supervision rather than labels, and this process can result in useful representation. The core hypothesis behind this idea is that solving these tasks need high-level semantic understanding of data, which forces self-supervised models to learn powerful representation.

Self-supervised learning is especially potential in video processing area since video is an information-intensive media that can provide plentiful contextual supervisory cues by nature. While various types of auxiliary strategies in video~\cite{Pulkit:agrawal2015learning,Xiaolong:wang2015unsupervised,Dinesh-et-al:jayaraman2016slow,Basura:fernando2017self} have shown impressive performance, there are two main \textit{limitations} of these works. \textbf{\textit{First}}, they commonly resort to a single task without accounting for complementarity among different task-specific features. Empirically, solving different tasks in video need different features and these features can complement each other to form a comprehensive understanding of video semantics. \textbf{\textit{Second}}, in order to achieve better performance, researchers tend to adopt deeper and wider models for representation embedding at the expense of high computational and memory cost. As video data in real-world workflows is of huge volume, efficiency issue must be addressed before practical application of classification approaches.

In this paper, we argue that heterogeneous video representations learnt from different auxiliary tasks in an ad-hoc fashion are not orthogonal among each other, which can be incorporated into a more robust feature-to-class semantics. In analogy to biological intelligence, humans can improve performance on a task via transferring knowledge learnt from other tasks and, intuitively, a general-purpose representation is strong enough to handle tasks in different scenarios. In light of above analysis, we propose to learn video representation for classification problem with a \textit{Divide and Conquer} manner: (1) instead of designing \textbf{one} versatile model for capturing discriminating features from different aspects simultaneously, we distill knowledge from \textbf{multiple} teacher models that are adept at specific aspects to the student model; (2) the student model is expected to be \textbf{lighter} since redundancy of knowledge from teachers is reduced after distillation. To this end, we propose a graph-based distillation framework, which bridges the advance of self-supervised learning and knowledge distillation for both exploiting complementaries among different self-supervised tasks and model compression. The main contributions of this paper are as follows:

(1) We propose logits graph ($ G_l $) to distill softened prediction knowledge of teachers at classifier level, where we formalize logits graph distillation as a multi-distribution joint matching problem and adopt Earth Mover (EM) distance as criteria to measure complementary gain flowing among the vertices of $ G_l $. (2) We propose representation graph ($ G_r $) to distill internal feature knowledge from pairwise ensembled representations yield by compact bilinear pooling, which tackles the challenge of heterogeneity among different features, as well as performs as an adaptation method to assist logits distillation.

Attributed to the above two distillation graphs, student model can incorporate complementary knowledge from multiple teachers to form a comprehensive video representation. Furthermore, distillation mechanism makes sure that student model works with fewer parameters than teachers, which has lower computational complexity and memory cost.

We conduct comprehensive experiments on 3 widely-used video classification datasets, which  validate that the proposed approach not only helps learn better video representation but also improve efficiency for video classification.

\section{Related Work}

\subsection{Self-supervised Learning}
Self-supervised learning is a recently introduced unsupervised paradigm, its key contribution is answering the question that how to effectively evaluate the performance of models trained without manual annotations. Typically, works in this area design tasks which are not directly concerned, such ``auxiliary'' tasks are difficult enough to ensure models can learn high-level representations. Nowadays, various self-supervised methods have been studied in computer vision, owing to image/video can provide rich structural information for developing auxiliary tasks.

Previous auxiliary tasks in single image domain involve asking networks to inpaint images with large missing regions~\cite{Deepak:pathak2016context}, colorize grayscale images~\cite{Richard:zhang2016colorful,Richard:Zhang_2017_CVPR} and solve jigsaw puzzles by predicting relative position of patches~\cite{Carl:doersch2015unsupervised,Mehdi:noroozi2016unsupervised}, \textit{etc}. Compared to images, videos contain more abundant spatiotemporal information to formulate self-supervision. For example, temporal continuity among frames can be leveraged to build a sequence order verification or sorting task~\cite{Ishan:misra2016shuffle,Basura:fernando2017self,Hsin-Ying:lee2017unsupervised}. Wang et al.~\shortcite{Xiaolong:wang2015unsupervised} finds corresponding patch pairs via visual tracking and use the constraint that similarity of matching pairs is larger than that of random pairs for training guidance. Analogously, Dinesh et al.~\shortcite{Dinesh-et-al:jayaraman2016slow} proposes temporally close frames should have similar features, as well as features change over time should be smooth. In addition, as shown in~\cite{Pulkit:agrawal2015learning}, ego-motion can also be utilized as meaningful supervision for visual representation learning. Our work is based on four self-supervised approaches in video and details are discussed in section \ref{sec:self-tasks}.

\subsection{Knowledge Distillation}
Knowledge distillation (KD)~\cite{Geoffrey:hinton2015distilling} aims to utilize the \textit{logits}, i.e., pre-softmax activations of trained classifiers (i.e., teacher models), to form softened probabilities that convey information of intra- and inter-class similarities. These extra supervisions (aka soft targets) can be combined with original one-hot labels (aka hard targets) to guide a lighter model (i.e., student model) to learn a more generalizable representation. After~\cite{Geoffrey:hinton2015distilling}, a surge of variants emerge to mine different forms of knowledge implied in teacher models. FitNets~\cite{Romero:fitnets} treats internal feature maps of ConvNet as hints that can provide explanation of how a teacher solves specific problems. Sergey et al.~\shortcite{Sergey:zagoruyko2016paying} propose learning to mimic the attention maps of teacher model can be helpful. Besides knowledge that teacher learnt from single sample, the relationships across different samples are also treasurable for student training~\cite{Yuntao:chen2017darkrank}. In this paper, we distill knowledge of multiple teachers from both logits prediction and internal feature, denoted by \textit{logits distillation} and \textit{representation distillation} respectively.

Lopez-Paz et al.~\shortcite{David:lopez2015unifying} propose \textit{generalized distillation} (GD) to combine KD with privileged information~\cite{Vladimir:vapnik2015learning}. This technique distills knowledge from privileged information learnt by teacher model at training stage, aiming at boosting student model at test stage where supervision from teacher is totally absent. More recently,~\cite{Zelun:luo2017graph} considers multi-modal data as privileged information and extends GD paradigm to a graph-based form. In this paper, we advocate that self-supervised tasks could be powerful privileged information which can supply student model with expertise from multiple views of video semantics, and internal feature distillation should be leveraged to further assist softened logits distillation as an adaptation method.

\section{Self-supervised Tasks}\label{sec:self-tasks}
\begin{figure*}[htb]
	\begin{center}
		\includegraphics[height=6.6cm]{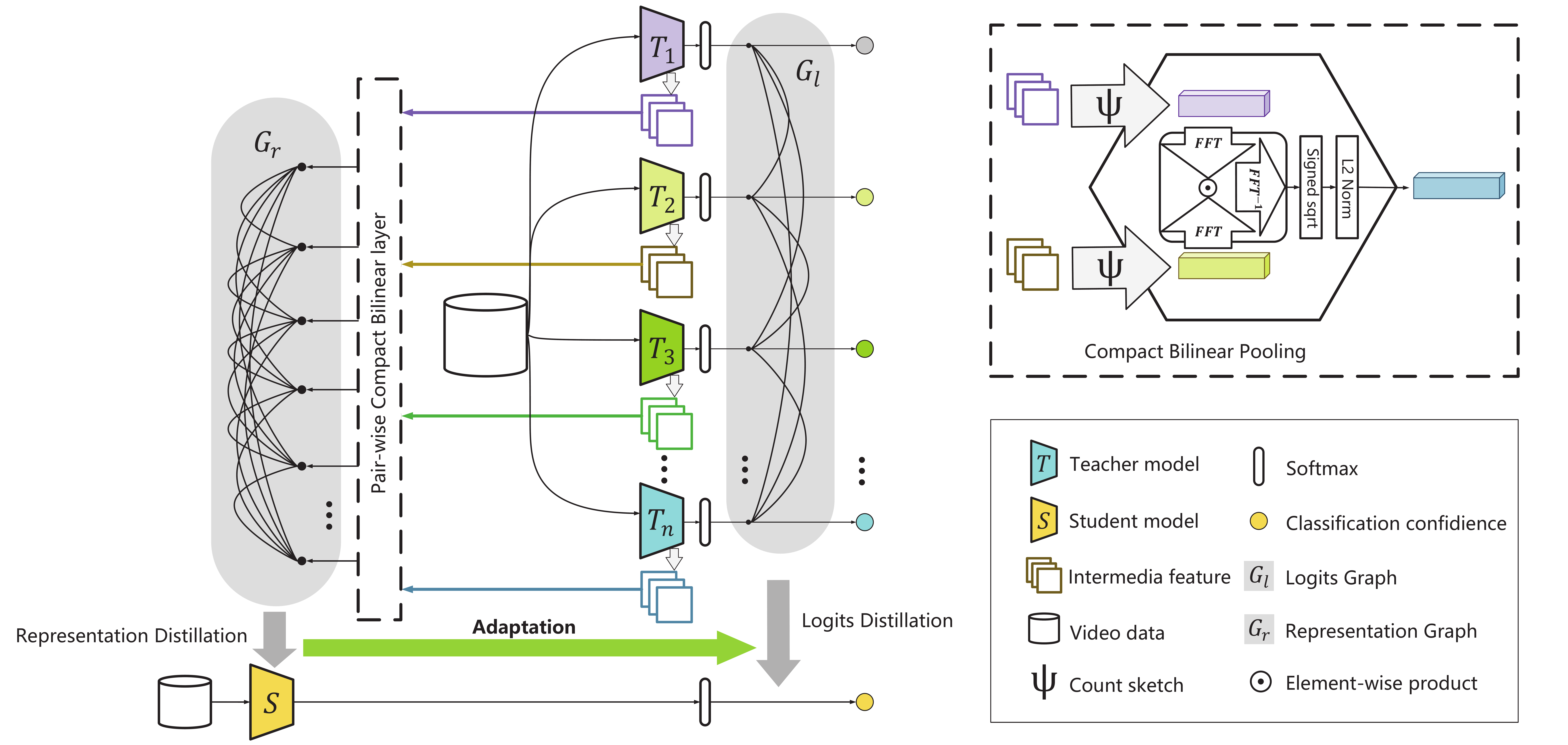}
	\end{center}
	\vspace{-1.3em}
	\caption{Architecture of the proposed framework.}
	\label{fig:framework}\vspace{-1.0em}
\end{figure*}
Under the principle of maximizing complementarity, we first implement four self-supervised methods in video, including frame sorting~\cite{Hsin-Ying:lee2017unsupervised}, learning from ego-motion~\cite{Pulkit:agrawal2015learning}, tracking~\cite{Xiaolong:wang2015unsupervised}, and learning from frame predicting that we design in this work inspired by~\cite{William:lotter2016deep}. For better readability, we denote them using abbreviations as $ \pi_\mathcal{S} $, $ \pi_\mathcal{M} $, $ \pi_\mathcal{T} $ and $ \pi_\mathcal{P} $, respectively. $ \pi_\mathcal{S} $ formulates the sequence sorting task as a multi-class classification problem. Taking symmetry of some actions (e.g., opening/closing a door) into account, there are $ n!/2 $ possible permutations for each $ n $-tuple of frames. $ \pi_\mathcal{M} $ captures camera transformation between two adjacent frames that the agent receives when it moves and trains model to determine whether two given frames are temporally close.  $ \pi_\mathcal{T} $ tracks similar patches in video and measures distance of them in representation space. $ \pi_\mathcal{P} $ predicts the subsequent frames of video, and we adopt a neuroscience inspired architecture named PredNet~\cite{William:lotter2016deep} as visual encoder. Self-supervised architectures and experimental details will be elaborated in section \ref{sec:5}.

We give the insights about the complementarity among these tasks as follows: (1) Self-supervised models can be mathematically divided into generative (e.g., $ \pi_\mathcal{P} $) and discriminative (e.g., other three tasks) categories, and features extracted by them represent two complementary aspects of video understanding, like the imagination and judgment aspects in human understanding. (2) $ \pi_\mathcal{S} $ and $ \pi_\mathcal{P} $ force machine to capture temporal information in video, while $ \pi_\mathcal{M} $ and $ \pi_\mathcal{T} $ focus on identifying visual elements or their parts, in which spatial feature is more useful. (3) Video contains both local and holistic feature that conveyed by short- and long-term clips, respectively. In our case, $ \pi_\mathcal{M} $ and $ \pi_\mathcal{T} $ are adept at utilizing local information, while $ \pi_\mathcal{S} $ and $ \pi_\mathcal{P} $ are committed to capture holistic feature in video.

\section{Graph Distillation Framework}\label{sec:4}
The proposed framework is illustrated in Fig. \ref{fig:framework}. We distill knowledge of self-supervised teacher models from two perspectives, namely soft probability distribution and internal representation. In this section, we first formalize the problem, then elaborate two components of the proposal.

\subsection{Problem Formalization}
To formalize the problem, assume $ \mathcal{D}_{tr}=\{\left(x_i,y_i\right)\}_{i=1}^{|\mathcal{D}_{tr}|} $ and $ \mathcal{D}_{te}=\{\left(x_i,y_i\right)\}_{i=1}^{|\mathcal{D}_{te}|} $ denote the training set with $ |\mathcal{D}_{tr}| $ samples and the test set with $ |\mathcal{D}_{te}| $ samples, respectively. $ x_i\in\mathbb{R}^d $ is the video clip and $ y_i\in[1,c] $ represents the label for $ c $-class video classification problem. $ \Pi=\{\pi_\mathcal{S},\pi_\mathcal{M},\pi_\mathcal{T},\pi_\mathcal{P}\} $ is the set of self-supervised tasks in section \ref{sec:self-tasks} and $ \mathcal{F}_t=\{f_t^{(i)}\}_{i=1}^{|\mathcal{F}_t|} $ is the set of teacher functions learnt from $ \Pi $. Our goal is learning a lighter student function $ f_s^\star $ with aid of $ \mathcal{F}_t $ under the principle of empirical risk minization (ERM):
\begin{equation}
f_s^\star=\arg\min\limits_{f_s\in{\mathcal{F}_s}}\frac{1}{|\mathcal{D}_{te}|}\sum\limits_{i=1}^{|\mathcal{D}_{te}|}\mathbb{E}\big[\ell(f_s(x_i),y_i)\big]+\Omega(\parallel\!f_s\!\parallel)
\label{eq:objective}
\end{equation}
where $ \mathcal{F}_s $ is the function class of $ f_s:\mathbb{R}^d\rightarrow[1,c] $, $ \ell $ is the loss function and $ \Omega $ is the regularizer. Notably, $ f_t $ and $ f_s $ in this paper are deep convolutional neural networks unless otherwise specified (see section~\ref{sec:architecture}).

\subsection{Logits Graph Distillation}
In vanilla knowledge distillation, the loss in Eq.~\eqref{eq:objective} is composed with two parts:
\begin{equation}
\ell(f_s(x_i),y_i)=(1-\lambda)\ell_{h}(f_s(x_i),y_i)+\lambda\ell_{s}(f_s(x_i),q_i)
\label{eq:KDeq}
\end{equation}
where $ \ell_{h} $ denotes the loss stems from true one-hot labels (i.e., hard targets) and $ \ell_{s} $ denotes the imitation loss that comes from softened predictions (i.e., soft targets), $ \lambda\in(0,1) $ is the hyperparameter which balances $ \ell_{h} $ and $ \ell_{s} $. In practice, $ \ell_{h} $ is typically the cross entropy loss:
{\setlength\abovedisplayskip{0.1pt}
\setlength\belowdisplayskip{0.7pt}
\begin{equation}
\ell_h(f_s(x_i),y_i)=\sum\limits_{k=1}^{c}\mathbb{I}(k=y_i)\log\sigma(f_s(x_i),y_i)
\label{eq:crossentropy}
\end{equation}}where $ \mathbb{I} $ is the indicator function and $ \sigma $ is the softmax operation $ \sigma(z_i)=\frac{\exp z_i}{\sum_{k=1}^{c}\exp z_k} $. Softened prediction $ q_i $ in imitation loss of Eq.~\eqref{eq:KDeq} is defined as $ q_i=\sigma\big(f_t(x_i)/T\big) $ in which $ f_t(x_i) $ is the class-probability prediction on $ x_i $ produced by teacher $ f_t $, $ T $ is the temperature and a higher value for $ T $ produces a softer probability distribution over the classes.

Logits graph $ G_l $ extends unidirectional distillation above to distilling logits knowledge from multiple self-supervised tasks dynamicly. $ G_l $ performs as a directed graph in which each vertex $ v_m $ represents a self-supervised teacher and the edge $ e_{n\rightarrow m}\in[0,1] $ denotes the information weight from $ v_n $ to $ v_m $. Given the video sample $ x_i $, the corresponding total imitation loss in logits graph distillation is calculated by
{\setlength\abovedisplayskip{5.0pt}
\setlength\belowdisplayskip{3.9pt}
\begin{equation}
\ell_s(x_i,\mathcal{F}_t^{i})=\sum\limits_{v_m\in{{U}}}\sum\limits_{v_n\in\mathcal{V}(v_m)}e_{n\rightarrow m}\cdot \ell_{n\rightarrow m}^{logit}(x_i,\mathcal{F}_t^{i})
\label{eq:softenedloss}
\end{equation}}where $ \ell_{n\rightarrow m}^{logit} $ means the distillation loss flowing on edge $ e_{n\rightarrow m} $, $ {U} $ is the universe set that contains all vertices of $ G_l $ and $ \mathcal{V}(v_m) $ is the set of vertices that point to $ v_m $. $ \mathcal{F}_t^{i} $ represents the logits output of teachers on sample $ x_i $.
\begin{figure}[htb]
	\begin{center}
		\includegraphics[height=2.9cm]{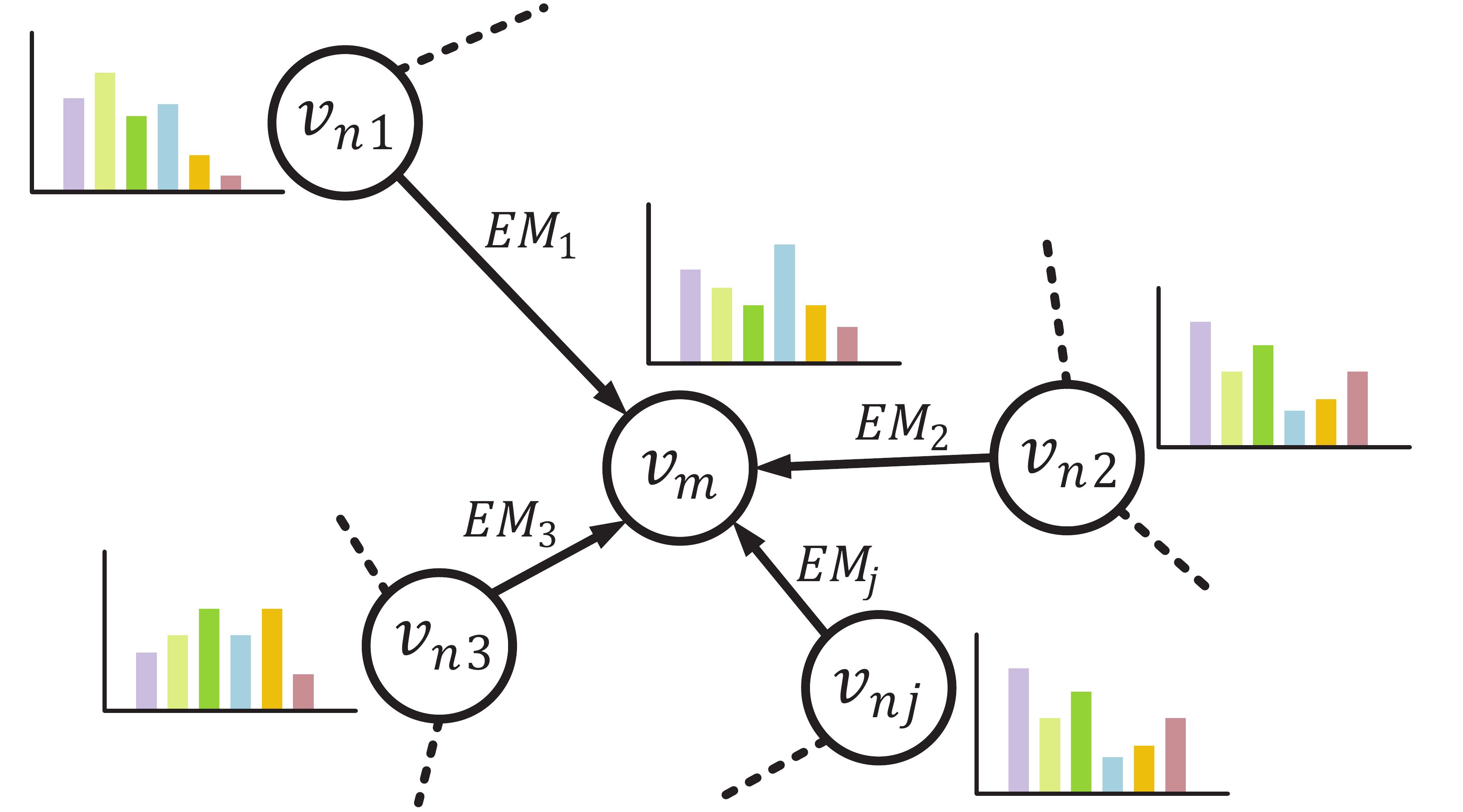}
	\end{center}
	\vspace{-0.9em}
	\caption{Illustration of logits graph distillation. EM denotes Earth Mover distance and the length of edges represents the information weight between adjacent vertices.}
	\label{fig:G_l}\vspace{-1.2em}
\end{figure}
As depicted in Fig.~\ref{fig:G_l}, we treat distillation in $ G_l $ as a multi-distribution matching problem, where teachers dynamicly adjust their logits distribution according to complementary gain (CG) received from their neighbors, and CG is based on the imitation feedback of student when it distills knowledge from different teachers. Take single teacher-student pair for example, we use Earth Mover (EM) distance to measure $ \ell^{logit}(x_i,f_t^{(k)}) $ as:
\begin{equation}
\ell^{logit}(x_i,f_t^{(k)})\!=\!\!\inf\limits_{\gamma\sim\Phi(\mathbb{P}_t^{(k)}\!,\mathbb{P}_s)}\!\!\mathbb{E}_{(\mu,\eta)\sim\gamma}\big(\!\!\parallel\!\mu(x_i)\!-\!\eta(x_i)\!\parallel\!\!\big)
\end{equation}
\begin{equation}
\mu(x_i)=\sigma(f_t^{(k)}(x_i)/T_k), \eta(x_i)=\sigma(f_s(x_i))
\end{equation}where $ \mathbb{P}_t^{(k)} $ and $ \mathbb{P}_s $ denote the space of probability belong to $ f_t^{(k)} $ and $ f_s $, respectively. $ \Phi(\mathbb{P}_t^{(k)},\mathbb{P}_s) $ represents the set of all joint distributions of $ \gamma(\mu,\eta) $ whose marginals are respectively $ \mathbb{P}_t^{(k)} $ and $ \mathbb{P}_s $. $ \mu(x_i) $ distills softened logits from $ f_t^{(k)} $ with temperature $ T_k $, where $ k $ means applying different temperature for different teachers. $ \eta(x_i) $ is output probability of the student network. We implement $ G_l $ using adjacency matrix where element  $ G_l[m][n] $ equals to $ e_{m\rightarrow n} $, and distillation is trained by minimizing the sum of Eq.~\eqref{eq:crossentropy} and Eq.~\eqref{eq:softenedloss} over all training samples iteratively.

The closest work to our $ G_l $ is multi-modal graph distillation (MMGD)~\cite{Zelun:luo2017graph} which distills knowledge from multi-model data, and there are mainly two differences between them. (1) We formalize logits distillation from multiple teachers as a multi-distribution joint matching problem.
Instead of cosine distance, we apply EM distance to measure the logits distribution discrepancy between teacher and student. Mathematically, teacher models are devoted to mapping visual characteristics of various video samples in logits distribution supported by low dimensional manifolds, where EM distance is proved to be more useful~\cite{Martin:arjovsky2017wasserstein}. (2) Moreover, we propose a novel feature distillation graph $ G_r $ to assist this distribution matching process (section~\ref{sec:G_r}) and experimental results indicate that our $ G_l $ can benefit much more from $ G_r $ than MMGD.

\subsection{Representation Graph Distillation}\label{sec:G_r}

Besides logits distribution, we hypothesize that intermediate feature of different teacher models can also be leveraged via graph-based distillation. The core difficulty stems from the heterogeneous nature of these task-specific features. To tackle this challenge, we propose to pairwise ensemble the original features via compact bilinear pooling~\cite{Yang:gao2016compact} and adopt the bilinear features as vertices of representation distillation graph $ G_r $, which is the second component in Fig~\ref{fig:framework}. There exist three insights behind this idea: (1) It is reasonable that the internal features of ConvNets reflect certain activation and attention patterns of neurons, and bilinear pooling allows all elements of heterogeneous feature vectors to interact with each other in a multiplicative fashion, which captures the salient information with more complementarity. (2) In the view of feature domain adaptation, bilinear pooling maps original features to a smoother representation space, where their distributions are more homologous and discrepancy among them are easier to measure. (3) Furthermore, since the high level prediction distribution of models is based on their internal representation, aligning at feature level also performs as a powerful adaptation method which assists distillation learning in $ G_l $.

Assume $ \mathcal{R}\in\mathbb{R}^{|C\times S|} $ denotes the feature vector with $ C $ channels and spatial size $ S $, where $ S=H\times W $. Each vertex $ V_k $ of $ G_r $ is yield by invoking compact bilinear pooling on feature pairs $ (\mathcal{R}_m,\mathcal{R}_n) $:
{\setlength\abovedisplayskip{3.9pt}
\setlength\belowdisplayskip{3.9pt}
\begin{multline}
\begin{aligned}
\!\!\!\Psi(\mathcal{R}_m\!\!\otimes\!\mathcal{R}_n,h,s)\!&=\!\Psi(\mathcal{R}_m,h,s)\ast\Psi(\mathcal{R}_n,h,s) \\
&=\!F\!FT^{-1}\!(F\!FT(\Psi_m)\!\odot\!F\!FT(\Psi_n))
\end{aligned}
\label{eq:compact}
\end{multline}}where $ \Psi:\mathbb{R}^{|C\times S|}\rightarrow\mathbb{R}^{e} $ represents Count Sketch projection function~\cite{Moses:charikar2002finding} and $ e\ll|C\times S| $ as $ \Psi $ projects the outer product to a lower dimensional space. $ \otimes $ and $ \odot $ are outer product and element-wise product, respectively. $ h\in\{1,\cdots\!,e\}^{|C\times S|} $ and $ s\in\{-1,1\}^{|C\times S|} $ are two randomly initialized vectors for invocation of $ \Psi $. $ \ast $ means convolution and Eq.~\eqref{eq:compact} utilizes the theorem that element-wise product in the frequency domain equals to convolution in time domain, which can yield a more compact feature vector. Then the bilinear feature $ \Psi $ is passed through a signed squareroot ($ z=sign(\Psi)\sqrt{|\Psi|} $) and $ \ell_2 $ normalization ($ V_k=z/\!\parallel\! z\!\parallel_2 $), where $ \forall k\in\{1,\mathcal{C}_{|\Pi|}^2\} $ and combinatorics $ \mathcal{C}_{|\Pi|}^2 $ is the vertex number of $ G_r $, for instance, there will be 6 vertices in $ G_r $ after pairwise ensemble of 4 teacher features for each video sample.

Similar to $ G_l $, $ G_r $ is defined as a adjacency matrix but the edge of $ G_r $ is a vector $ E\in\mathbb{R}^{1\times b} $ rather than a scalar, where $ b $ is the dimensention of vertex $ V $. We distill the bilinear feature with temperature $ T_k $ as softened representation distribution
{\setlength\abovedisplayskip{4.9pt}
\setlength\belowdisplayskip{3.9pt}
\begin{equation}
\mathcal{D}_k^{soft}(x_i)=\sigma(V_k(x_i)/T_k)
\label{eq:soft_representation}
\end{equation}}and use Maximum Mean Discrepancy (MMD) to meause the distillation loss:
{\setlength\abovedisplayskip{0.1pt}
\setlength\belowdisplayskip{0.7pt}
\begin{multline}
\begin{aligned}
\!\!\!\ell_r(\mathcal{R}_s,\mathcal{D}_k^{soft})&\!=\frac{1}{C_s^2}\sum\limits_{p=1}^{C_s}\sum\limits_{p'=1}^{C_s}\mathcal{K}\big(\sigma(\mathcal{R}_s^{(p)}),\sigma(\mathcal{R}_s^{(p')})\big) \\
&\!+\frac{1}{C_k^2}\sum\limits_{q=1}^{C_k}\sum\limits_{q'=1}^{C_k}\mathcal{K}\big(\mathcal{D}_k^{(q)},\mathcal{D}_k^{(q')}\big) \\
&\!-\frac{2}{C_sC_k}\sum\limits_{p=1}^{C_s}\sum\limits_{q=1}^{C_k}\mathcal{K}\big(\sigma(\mathcal{R}_s^{(p)}),\mathcal{D}_k^{(q)}\big)
\end{aligned}
\label{eq:MMDloss}
\end{multline}}where $ \mathcal{R}_s $ denotes the feature map in a certain layer of student model and $ \mathcal{R}_s^{(p)} $ is the feature vector at channel $ p $. $ \mathcal{D}_k $ is the softened distribution of $ k $-th vertex in $ G_r $, $ C_s $ and $ C_k $ are channel numbers of student model and $ \mathcal{D}_k $, respectively. $ \mathcal{K} $ is kernel function which maps features to higher dimensional Reproducing Kernel Hilbert Space (RKHS) and we use Gaussian Kernel: $ \mathcal{K}(\bm{a},\bm{b})=\exp(-\frac{\parallel\bm{a}-\bm{b}\parallel_2^2}{2\sigma^2}) $ in practice. Note that, softmax operation in Eq.~\eqref{eq:MMDloss} ensures each sample has the same scale and it is unnecessary for $ \mathcal{D}_k $ as Eq.~\eqref{eq:soft_representation} have already performed it.

Let $ \mathcal{M}_{\ell_r} $ is the matrix whose element is $ \ell_r $ and $ \mathcal{T}\in\mathbb{R}^{1\times \mathcal{C}_{|\Pi|}^2} $ is the temperature vector, the total loss of $ G_r $ is calculated by
\begin{equation}
\mathcal{L}_r=\sum\sigma(G_r/\mathcal{T})\odot\mathcal{M}_{\ell_r}
\end{equation}

\section{Experiments}\label{sec:5}
In this work we use convensional transfer learning paradigm for evaluating the utility of student model on video classification. We first train the teacher models for privileged information learning on auxiliary tasks, then transfer the knowledge from teachers to student via the proposed graph distillation framework. Experimental details are described in the sequel.

\subsection{Architectures}\label{sec:architecture}
We first reimplement three self-supervised methods in the literature based on VGG-19~\cite{Karen:simonyan2014very}, for training teacher model from $ \pi_\mathcal{S} $, $ \pi_\mathcal{M} $ and $ \pi_\mathcal{T} $, respectively. Then we introduce PredNet~\cite{William:lotter2016deep} as a self-supervised model for learning from $ \pi_\mathcal{P} $. For notation convenience, we use $ T_\mathcal{S} $, $ T_\mathcal{M} $, $ T_\mathcal{T} $, and $ T_\mathcal{P} $ to denote the models for these tasks. All models are customized with a slight modification for meeting the demand in our experiments. In particular, we respectively extend channels to 96 and 128 for {\tt{conv1}} and {\tt{conv2}} in VGG-19, and change the filter size of {\tt{conv1}} to 11$ \times $11 as better performance have shown in practice. For $ \pi_\mathcal{S} $, $ \pi_\mathcal{M} $ and $ \pi_\mathcal{T} $, siamese-style architectures are conducted for pairwise feature extraction and base visual encoders share parameters. More specifically, $ T_\mathcal{S} $ concatenates convolution feature over pairs of all frames to be sorted in first {\tt{fc}} layer, followed by a classical {\tt{fc}} classifier. $ T_\mathcal{M} $ is composed with a base-CNN (BCNN) and a top-CNN (TCNN), in which TCNN take the output pair of BCNN for transform analysis between two frames. $ T_\mathcal{T} $ is a siamese-triplet network in order to judge the similarity of patches in a triple. $ T_\mathcal{P} $ is a ConvLSTM model where each layer consists of representation module ($ R $), target module ($ A $), prediction module ($ \hat{A} $) and error term ($ E $). We conduct a 4-layers version of it and use the output of $ R $ in top layer as representation. For the sake of model compression, we choose AlexNet~\cite{Alex:krizhevsky2012imagenet} as student model $ S $ for video classification task. 

\subsection{Implementation Details}
All models are implemented using PyTorch\footnote{\tt http://pytorch.org/}. For the self-supervised model training, we basically follow the settings in original papers and we encourage the reader to check them for more details. For the graph distillation training, we train the model for 350 epochs using the Adam optimizer with momentum 0.9. The batch size is set to 128 and we initialize the learning rate to 0.01 and decay it every 50 epochs by a factor of 0.5. Trade-off hyperparameter $ \lambda $ is set to 0.6.

Another vital issue in practice is how to choose the layer of different networks for feature distillation in $ G_r $. In other words, we expect to quantify the transferability of features from each layer of teacher models. Inspired by~\cite{Jason:yosinski2014transferable}, we measure 1) the specialization of layer neurons to original self-supervised tasks at the expense of performance on the classification task and 2) the optimization difficulties related to splitting teacher networks between co-adapted neurons, and we finally choose {\tt conv4} of both teachers (VGG-19 and PredNet) and student (AlexNet) as the distillation layer, whose feature maps have 256 out channels.

\subsection{Datasets}
Datasets in our experiments contain two branches, i.e., auxiliary datasets for self-supervised learning and target datasets for video classification. For the former, we use the same datasets as original works. For the latter, we evaluate our framework on three popular datasets, namely UCF101~\cite{Khurram:soomro2012ucf101}, HMDB51~\cite{Hilde:kuehne2013hmdb51} and CCV~\cite{Yu-Gang:jiang2011consumer}, which contain 13k, 6.7k and 9.3k videos with 101, 51 and 20 categories, respectively.

\subsection{Results}\label{sec:results}
Table~\ref{tab:1} compares our framework to several state-of-the-art unsupervised methods for video classification. We report the results averaged over 3 splits on both UCF101 and HMDB51. Obviously, our approach outperforms the previous methods by a large margin, and more importantly, it beats its ImageNet-supervised counterpart ($ AlexNet_{pre} $) \textbf{for the first time}. The results indicate that (1) knowledge learnt from different self-supervised tasks is complementary and it can be transferred via graph distillation to form a more comprehensive video semantics. (2) Knowledge from models which trained on ImageNet in a supervised fashion are insufficient for capturing all the details of video, leading to a suboptimal performance on classification task.


\begin{table}[htbp]
	\centering
	\vspace{-0.46em}
	\setlength{\abovecaptionskip}{3.8pt}%
	\caption{Comparing with other state-of-the-art unsupervised methods for video classification. Average \% accuracy for UCF101 and HMDB51, mean average precision for CCV. $ AlexNet_{ran} $ means AlexNet trained from scratch with randomly initialized weights.}
	\label{tab:1}
	\begin{tabular}{lccc}
		\toprule
		Methods & UCF101 & HMDB51 & CCV \\
		\midrule
		$ AlexNet_{ran} $                                                & 47.8          & 16.3           & 39.2 \\
		$ AlexNet_{pre} $                                                & 66.9          & 28.0           & 60.8 \\
		Wang \textit{et al.}~\shortcite{Xiaolong:wang2015unsupervised}   & 40.7          & 15.6           & 34.3 \\
		Misra \textit{et al.}~\shortcite{Ishan:misra2016shuffle}         & 50.9          & 19.8           & 44.7 \\
		Senthil \textit{et al.}~\shortcite{Senthil:purushwalkam2016pose} & 55.4          & 23.6           & 46.4 \\
		Lee \textit{et al.}~\shortcite{Hsin-Ying:lee2017unsupervised}    & 56.3          & 22.1           & 48.8 \\
		Basura \textit{et al.}~\shortcite{Basura:fernando2017self}       & 60.3          & 32.5           & 58.1 \\
		\midrule
		Ours ($ G_l $+$ G_r $)                                           & \textbf{68.9} & \textbf{35.1}  & \textbf{62.6} \\
		\bottomrule
	\end{tabular}
\vspace{-1.0em}
\end{table}

\subsubsection{Representation Learnt by Student}
\begin{figure}[htb]
	\begin{center}
		\includegraphics[height=5.3cm]{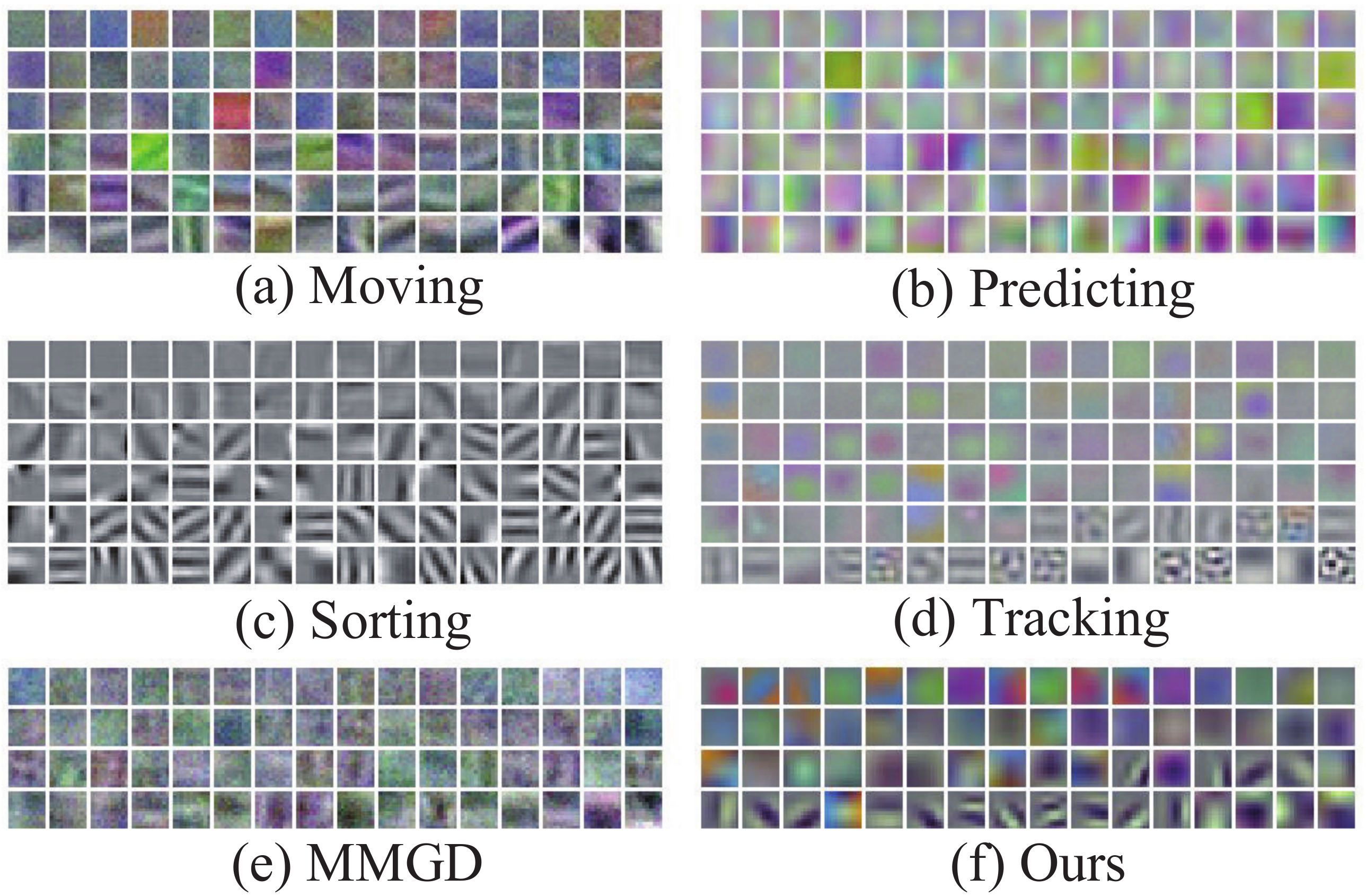}
	\end{center}
	\vspace{-1.2em}
	\caption{Visualization of {\tt{conv1}} filters learnt from self-supervised tasks and graph distillation. Filters in (a), (b), (c) and (d) are learnt from $ \pi_{\mathcal{M}} $, $ \pi_{\mathcal{P}} $, $ \pi_{\mathcal{S}} $ and $ \pi_{\mathcal{T}} $, respectively. The filters of $ \pi_{\mathcal{S}} $ are in grayscale since we use channel splitting for better performance. (e) and (f) show the filters of student trained on UCF101 via MMGD and our proposal, both of them adopt AlexNet as base network.}
	\label{fig:filters}\vspace{-1.0em}
\end{figure}
Qualitatively, we demonstrate the quality of the features learnt by student model through visualizing its low-level first layer filters ({\tt conv1}). As exhibited in Fig.~\ref{fig:filters}, there are both color detectors and edge detectors in the {\tt conv1} filters of our distilled model. Its filters are sharper and of more varieties than its counterpart learnt from MMGD, and tend to be fully utilized when compared to the teachers.


\subsection{Ablation Studies}
We conduct comparison with several baseline methods to evaluate the effectiveness of each component in our proposal. After fine-tuning the self-supervised teacher models for video classification individually, we fuse their confidence scores (model fusion). We also feed the spliced feature of them to a linear classifier (model ensemble). For the conventional knowledge distillation (KD) baseline, we average the confidence scores of teacher models that equipped with KD. $ G_l $ and $ G_r $ are respectively compared with MMGD and single feature distillation (uniform version of feature distillation). Moreover, in order to verify the mutual promotion between $ G_l $ and $ G_r $, we further compare the performance of two combination, i.e., MMGD+$ G_r $ and $ G_l $+$ G_r $ (ours).
\begin{table}[htbp]
	\centering
	\vspace{-0.40em}
	\setlength{\abovecaptionskip}{3pt}%
	\caption{Comparison with baseline methods on three vdieo datasets.}
	\label{tab:baselineTab}
	\begin{tabular}{lccc}
		\toprule
		Methods 						& UCF101 & HMDB51 & CCV \\
		\midrule
		Models fusion 					& 57.9 & 19.4 & 45.2 \\
		Models ensemble 					& 58.3 & 23.5 & 47.1 \\
		KD (uniform) 					& 60.1 & 24.3 & 48.1 \\
		\midrule
		Logits graph (MMGD) 			& 62.5 & 27.2 & 49.7 \\
		Logits graph ($ G_l $) 			& \textbf{65.3} & \textbf{30.9} & \textbf{52.6} \\
		\midrule
		Feature distill (uniform) 		& 61.7 & 28.4 & 56.3 \\
		Feature graph distill ($ G_r $) & \textbf{66.4} & \textbf{33.4} & \textbf{58.0} \\
		\midrule
		MMGD+$ G_r $ 					& 67.0 & 34.2 & 58.9 \\
		$ G_l $+$ G_r $ 				& \textbf{68.9} & \textbf{35.1} & \textbf{62.6} \\
		\bottomrule
	\end{tabular}
\vspace{-0.6em}
\end{table}

\begin{figure}[htb]
	\vspace{-0.5em}
	\begin{center}
		\includegraphics[height=3.55cm]{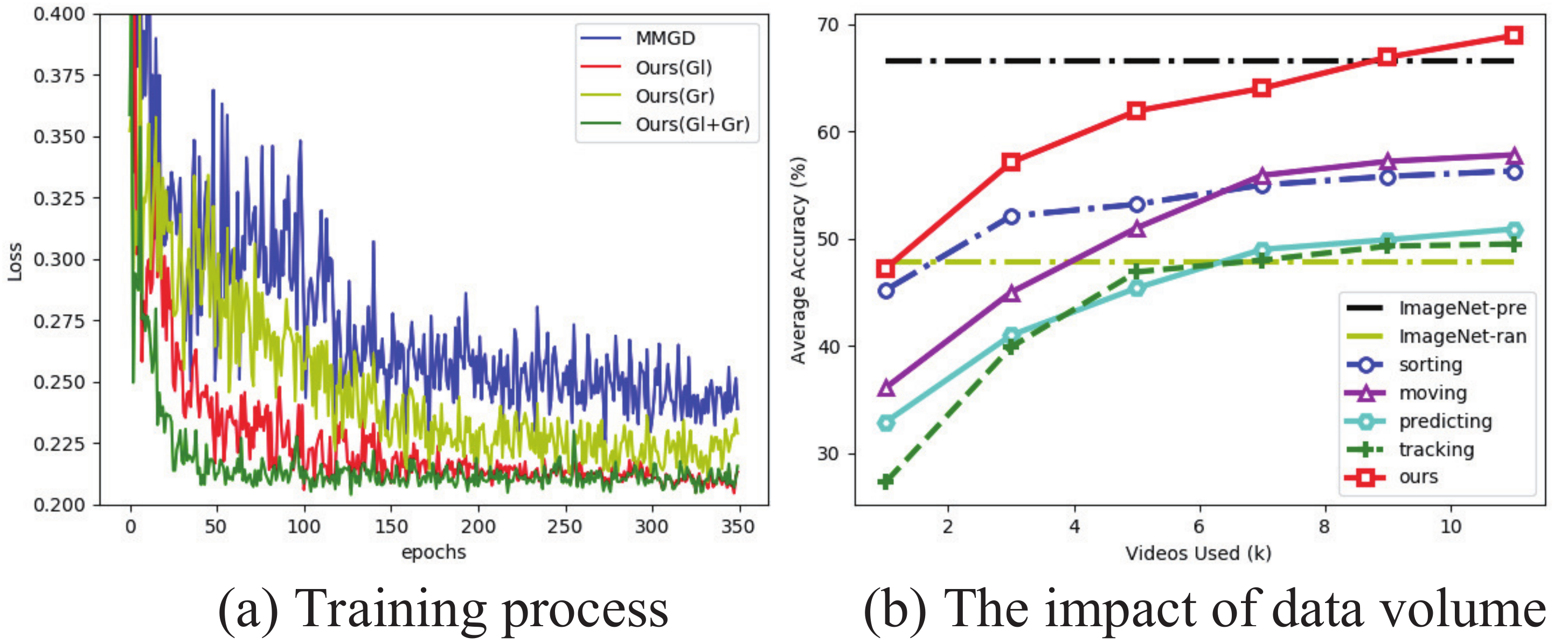}
	\end{center}
	\vspace{-1.0em}
	\caption{Illustration of training process and the impact of video data volume for the classification performance on UCF101.}
	\label{fig:twofig}\vspace{-1.2em}
\end{figure}

As shown in Table~\ref{tab:baselineTab}, graph-based distillation methods outperform the ordinary solutions that in the first three lines. The comparison between row 4 and row 5 suggests that our formalization of logits graph distillation is more effective than MMGD. The comparison between row 6 and row 7 reveals $ G_r $ helps model to learn the complementarity of different self-supervised tasks. Results in the last two lines verify that $ G_l $ benefits much more from $ G_r $ than MMGD since $ G_r $ performs as a powerful adaptation method to assist the probability distribution matching. Moreover, the training process of distillation shown in Fig.~\ref{fig:twofig} (a) suggests that $ G_r $ can boost $ G_l $ not only in accuracy but also in training efficiency, which further verify the effectiveness of the two components.
\subsection{Comparison with Supervised Methods}
Although our approach has not yet yielded video representations as powerful as those learned by supervised classification methods, it shows the advance of integrating complementary video semantics from different self-supervised models. One of the most significant advantages of our approach is that it can utilize massive video data on the web economically, and as displayed in Fig.~\ref{fig:twofig} (b), the performance of distilled model increases steadily as the training data for self-supervised tasks grows. We believe learning video representation in such \textit{divide and conquer} manner is potential as it can integrate complementary semantic from various models at expert level.

\subsection{Model Compression}
Attributed to the knowledge distillation, our proposal dramatically reduces the number of model parameters and inference time, since top performing models for video classification has a much higher computational and memory cost than AlexNet. Moreover, student mdoel is much lighter than teacher models. 

\section{Conclusion}
In this paper, we have proposed a graph-based distillation framework for representation learning in video classification, which can leverage the complementarity of different video semantics arising from multiple self-supervised tasks. Our proposal distills both logits distribution knowledge and internal feature knowledge from teacher models. We formulate the former as a multi-distribution joint matching problem which is proved to be more effective than previous methods. For the latter, we propose to pairwise ensemble the different features via compact bilinear pooling, and this solution works as not only a heterogeneity eliminator for feature distillation but also a powerful adaptation method to further assist the logits distillation. Furthermore, two graphs ensure a lighter student can learn to use its full capacity for video representation learning as the redundancy of knowledge from teacher models is reduced by distillation.

\section*{Acknowledgments}
This work was supported by National Natural Science Foundation of China under Grant 61771025 and Grant 61532005.

\bibliographystyle{named}
\bibliography{ijcai18}

\end{document}